\documentclass[runningheads]{llncs}

\usepackage[year=2026]{eccv}
 
\usepackage{eccvabbrv}

\usepackage{graphicx}
\usepackage{booktabs}

\usepackage{tcolorbox}
\usepackage{tabularx}
\usepackage{subcaption}

\usepackage{graphicx}
\usepackage{booktabs}

\usepackage{tcolorbox}
\usepackage{tabularx}

\usepackage{algorithm}
\usepackage{algpseudocode}

\usepackage[accsupp]{axessibility}  

\usepackage{xcolor}

\usepackage{hyperref}

\usepackage[accsupp]{axessibility}   

\usepackage{xcolor}

\usepackage{hyperref}

\newcommand{\name}{\textbf{LATERN}}

\begin{document}

\title{{\name}:   Test-Time  Context-Aware Explainable Video Anomaly Detection}

\author{Mitchell Piehl\inst{1} \and
Muchao Ye\inst{1}\thanks{Corresponding author.}}

\institute{The University of Iowa, Iowa City IA 52245, USA \\
\email{mpiehl@uiowa.edu, muchao-ye@uiowa.edu}\\
}

\maketitle

\begin{abstract}
    Vision-language models (VLMs) have recently emerged as a promising paradigm for video anomaly detection (VAD) due to their strong visual reasoning ability and natural language-based explainability. In this paper, we aim to address a key limitation of such pipelines, which perform segment-level inference independently owing to token constraints and reason without structured temporal context, allowing VLMs to interpret anomalies as deviations from evolving video dynamics rather than producing fragmented predictions and explanations.
    To specify, we propose a context-aware framework named {\name}, which reformulates VAD as a temporal evidence aggregation process. {\name} consists of two complementary modules: Context-Aware Anomaly Scoring (CEA) and Recursive Evidence Aggregation (REA). CEA introduces a novel image-grounded memory mechanism, which selectively chooses historical content via frame diversity and visual-textual alignment as expanded context to help generate reliable anomaly scores.  
    Building upon these scores, REA performs recursive temporal aggregation to identify coherent anomaly intervals and produce event-level decisions and explanations grounded in visual–textual evidence. Extensive experiments on challenging benchmarks, including UCF-Crime and XD-Violence, show that {\name} enhances detection accuracy and explanation consistency for frozen VLMs during test time, while generating temporally coherent and semantically grounded event-level explanations.
\end{abstract}

\section{Introduction}
\label{sec:intro}

\begin{figure}[tb]
    \centering
    \includegraphics[width=1.0\linewidth]{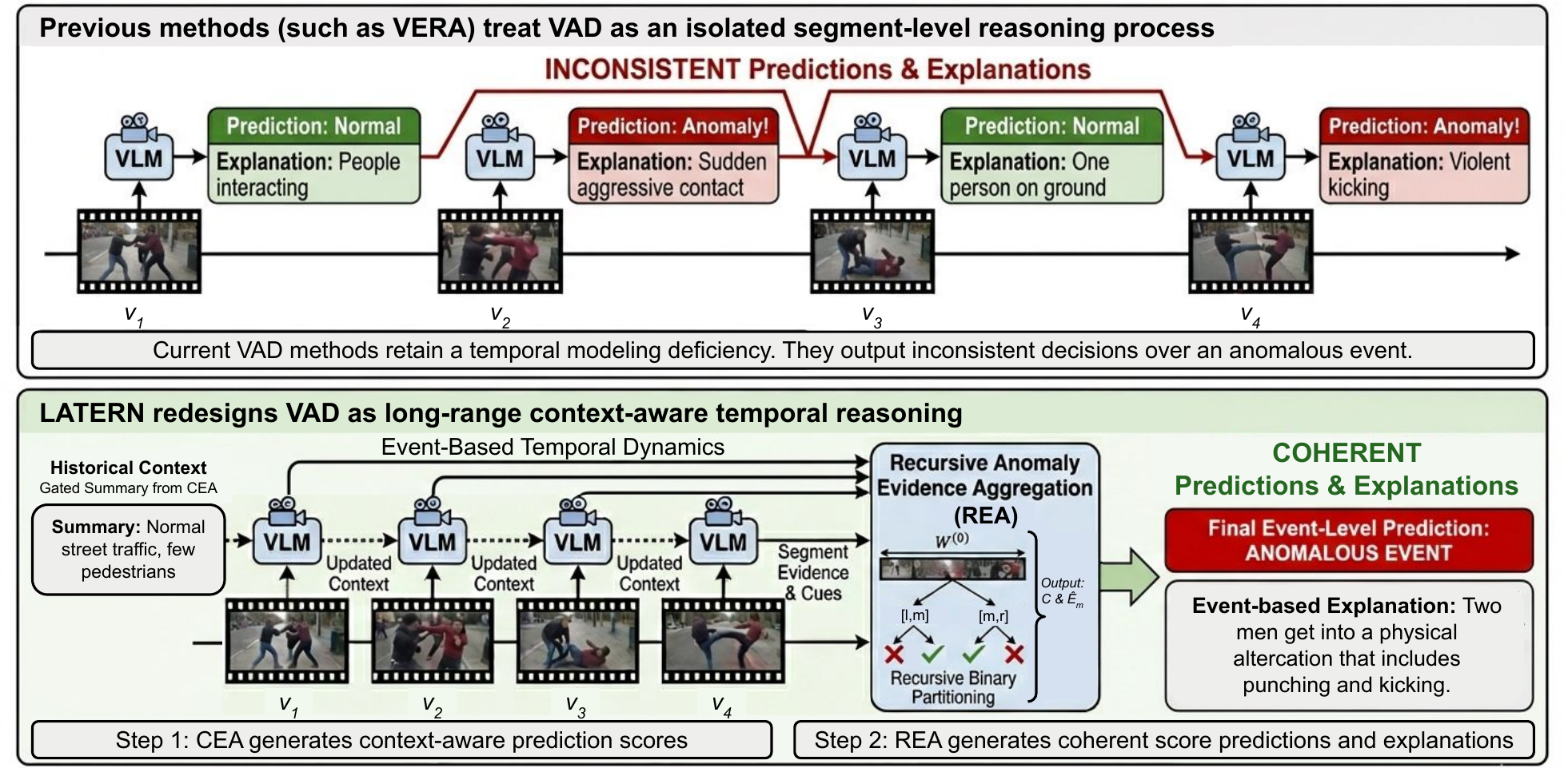}
    \caption{Illustration of the limitations of previous VAD methods using isolated segment-level reasoning compared to our proposed event-level reasoning pipeline using historical context and anomalous event aggregation.}
    \label{fig:WeaknessAddressedFigure}
\end{figure}

Video anomaly detection (VAD) has received significant interest within the AI community due to its tremendous benefit in achieving automated decision-making for safety-critical applications, including video surveillance~\cite{ramachandra2020survey}, medical diagnosis~\cite{fernando2021deep}, and autonomous driving~\cite{yao2022dota}. 
An ideal VAD system is expected to meet requirements on \textbf{accuracy} and \textbf{explainability}: to  correctly localize the occurrence of video anomalies and generate truthful explanations for users.
To achieve this vision, the research community has switched to using vision-language models (VLMs)~\cite{hurst2024gpt,Qwen2VL,chen2024internvl} for VAD because they can perceive visual inputs, follow textual instructions, and generate verbalized prediction results.

Existing studies have developed successful model adaptation strategies for VLMs in VAD, e.g., prompt engineering~\cite{ye2024vera,zanella2024harnessing,yang2024follow} during test time or fine-tuning~\cite{zhang2024holmes,zhang2025holmes,huangvad} in training. \textbf{However, they retain a temporal modeling deficiency in inference  when the model is adapted}: most current VLM-based methods~\cite{ye2024vera,zanella2024harnessing,yang2024follow, lee2025flashback} map each video segment independently to explanations and numerical prediction scores in the language space for VAD because processing long untrimmed videos would exceed token limits.
As shown in Fig.~\ref{fig:WeaknessAddressedFigure}, this leads to a fundamental limitation in achieving accurate detection and explanations for existing methods: (1) Without global temporal context, VLMs may yield fragmented or partial anomaly assessments, leading to fluctuating anomaly scores (e.g. inconsistent anomaly scores over a long period covering a single anomalous event like fighting) and suboptimal detection performance. Additionally, (2) explanations for anomalous segments turn unreliable because explanations are generated for every given segment and can usually be inconsistent with neighboring segments (e.g., a single fight over a long period may be described in many different variations). To bridge this gap, we investigate the following fundamental yet largely unexplored research question:

\noindent \emph{How can long-range video context be incorporated into VLM-based VAD to enable accurate and explainable reasoning during test time?}

We answer this research question by a principled framework that explicitly models \textbf{temporal aggregation of anomaly evidence during test time} for VLMs in VAD. The underlying intuition of our design is that anomalies are inherently context-dependent and must be interpreted relative to the global temporal dynamics of a video. To this end, we propose a novel framework named {\name}, redesigning VLM-based VAD as a \underline{l}ong-range context-\underline{a}ware \underline{te}mporal \underline{r}easo\underline{n}ing process. {\name} includes two important modules: Context-Aware Anomaly Scoring (\textbf{CEA}) and Recursive Evidence Aggregation (\textbf{REA}). Firstly, CEA introduces an \underline{image-grounded memory} module that accumulates long-range past segments and selectively aggregates informative ones as temporal evidence based on visual diversity as the extended context to generate historical summary during inference. In addition, CEA includes a gate mechanism in the memory based on visual–textual alignment to prevent inaccurate summaries from harming anomaly reasoning. This module ensures that VLMs perform anomaly reasoning in the current segment relative to a structured representation of prior video content. Secondly, the context-aware anomaly scores are used by REA  to recursively partition the video into multi-scale anomaly window proposals and getting coherent anomaly regions via  \underline{structured evidence accumulation}. Importantly, REA produces event-based explanations grounded in visual–textual evidence for each anomalous window. Thus, this module enables {\name} 
generate stable predictions and temporally consistent explanations. By addressing context deficiency during test time and emphasizing temporal consistency in predictions, {\name} transforms VLM-based VAD from isolated segment reasoning into structured, context-aware event understanding.

Accordingly, we make the following key contributions:
\begin{itemize}
    \item To our knowledge, we are the first to introduce a context-aware principle for VLMs in VAD to   address the key limitation of the existing paradigm in which VLMs conduct anomaly reasoning segment by segment  with limited context. The proposed {\name} framework helps VLM reason over long range temporally during test time, mitigating fragmented predictions and producing coherent anomaly localization in challenging real-world videos.

    \item Computationally, {\name} consists of two complementary modules (CEA and REA) to achieve this. CEA introduces an image-grounded memory with diversity-aware selection and visual–textual alignment gating to produce reliable context-aware anomaly scores. Building upon these scores, REA performs hierarchical temporal evidence aggregation to identify coherent anomaly intervals and generate event-level explanations grounded in visual–textual evidence. Together, the two modules transform vanilla segment-level inference into structured, temporally consistent anomaly reasoning.

    \item  Extensive experiments on large-scale VAD benchmarks including UCF-Crime and XD-Violence show that {\name} generates coherent event-level explanations grounded in anomaly evidence and significantly enhances detection performance with frozen VLM backbones during test time.

\end{itemize}

\section{Related Work}
\noindent\emph{Video Anomaly Detection.} Detecting video anomalies is inherently challenging due to the ambiguous and context-dependent nature of what constitutes an anomaly compared to image anomalies~\cite{zhu2025fine}. Prior to VLMs, commonly used are task-specific deep neural networks trained with weakly supervised classification tasks~\cite{sultani2018real, wu2024vadclip, tian2021weakly} or unsupervised frame reconstruction tasks~\cite{liu2018future,ye2019anopcn,lu2013abnormal}. However, they can only output numerical results and largely overlook the explainability aspect. To address this limitation, the recent line of literature introduces VLMs to generate predictions and explanations together. 
Prompt engineering~\cite{zanella2024harnessing,ye2024vera} or instruction tuning~\cite{zhang2024holmes,lv2024video,yang2024follow,tang2024hawk} is adopted to make VLM responsive to the instruction for VAD. 
However, existing VLM-based methods typically conduct inference at the level of isolated temporal segments, as processing long untrimmed videos would exceed token limits. Consequently, such segment-level reasoning leads to fragmented predictions and context-limited explanations, especially for anomalies that can only be properly interpreted with broader temporal context. In contrast, our framework {\name} explicitly addresses this limitation by enabling structured temporal evidence aggregation, thereby bridging the gap between context-aware anomaly reasoning and VLM-based explainable VAD.

\noindent\emph{Memory for LLMs/VLMs.}
Recent advances in agentic memory systems, such as A-Mem~\cite{xu2025mem} and others ~\cite{chhikara2025mem0, packer2023memgpt, maharana2024evaluating, rasmussen2025zep}, demonstrate the effectiveness of textual memory for long-horizon reasoning. However, these approaches operate purely in the textual embedding space and lack direct access to visual evidence, potentially obscuring fine-grained appearance cues that are critical for VAD. Meanwhile, memory mechanisms proposed for general video understanding (e.g., StreamForest~\cite{zengstreamforest}) focus on holistic temporal knowledge accumulation for recognition or summarization, rather than modeling deviation-sensitive temporal dynamics \cite{shen2024longvu, chen2024videollm, song2024moviechat, zhang2024flash, zhang2024long}. As a result, existing memory paradigms are not well suited for VAD. In contrast, {\name} fills this gap with a memory design in CEA which is image-grounded, selectively updated, and explicitly validated through cross-modal alignment, enabling deviation-aware context modeling for VAD.

\noindent\emph{Event-Level VAD Explanation.}
Recent VLM-based VAD approaches such as Holmes-VAU~\cite{zhang2025holmes} and CUVA~\cite{cuva2024} have demonstrated their ability to generate event-level anomaly descriptions. However, this capability is primarily achieved through  fine tuning of the deploying VLMs, which adapts model parameters to the target task. In contrast, our framework operates in a fully frozen VLM setting and does not rely on additional training, 
enabling coherent event-based explanation directly from test-time reasoning through structured temporal evidence aggregation. This distinction highlights a fundamentally different pathway toward explainable VAD, emphasizing test-time temporal reasoning.

\section{The {\name} Framework}
\begin{figure}[tb]
    \centering
    \includegraphics[width=1.0\linewidth]{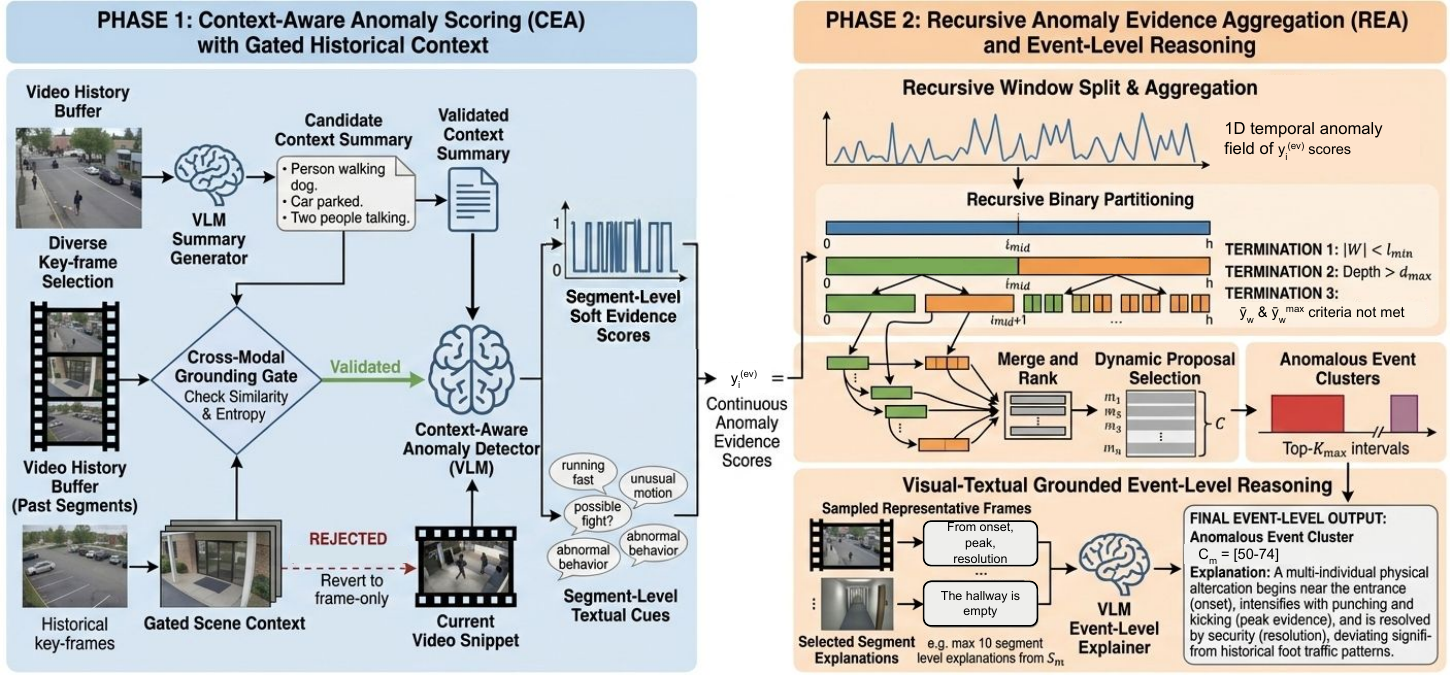}
    \caption{The proposed \name \space framework. Phase 1 (CEA) constructs and validates historical summaries for context-aware VLM inference. Phase 2 (REA) aggregates noisy, segment-level evidence via recursive binary partitioning temporally to identify coherent anomaly clusters, followed by multi-modal event-level explanation generation.}
    \label{fig:MainFigure}
\end{figure}

As shown in Fig.~\ref{fig:MainFigure}, the proposed {\name} framework will first expand temporal context and then recursively refine, aggregate anomaly evidence for precise localization and explanation during test time. In \S~\ref{sec:prob_formulate}, we first provide a problem statement for VAD. After that, in \S~\ref{sec:CEA}, we detail the \textbf{C}ontext-Awar\textbf{e} \textbf{A}nomaly Scoring mechanism (CEA)  in {\name} empowered by a new memory mechanism. Later, in \S~\ref{sec:RACA}, we present \textbf{R}ecursive Anomaly \textbf{E}vidence \textbf{A}ggregation (REA) in {\name},which progressively generate event-based temporal window to generate refined anomaly scores and natural language-based explanations.

\subsection{Problem Formulation: VLMs for VAD during Test Time} 
\label{sec:prob_formulate}
During inference in VAD, a video \( V = \{I_i\}_{i=1}^F\) with \( F \) frames is given, where \( I_i \) is the $i$-th frame. Frame-level ground truth \( Y = [y_1, \dots, y_F] \) is provided: \( y_i=1 \) if \( I_i \) is part of an anomalous event, and \( y_i=0 \) otherwise.  The employed VLM $f$ (base model) is expected to generate an anomaly score sequence $\hat{Y}=[\hat{y}_1,\cdots,\hat{y}_F]$ for all frames, where each $\hat{y}_i\in[0, 1] (1\le i \le F)$. However, 
since $F$ is huge,  existing works~\cite{ye2024vera,zanella2024harnessing,yang2024follow} adopt a coarse-to-fine anomaly scoring process: After dividing a given video \( V\)  into $h$ segments $\{v_{j}\}_{j=1}^{h}$ (see \S~\ref{sec:exp} for details), $f$ will first generate initial scores for each video segment \textbf{independently} as a set of  coarse-grained binary scores $[\tilde{y}_1,\cdots,\tilde{y}_h]$ by inputting a textual VAD-specific instruction $\theta$ and each $v_j$  to \( f \) to generate an explanation $E_j$  whether an anomaly exists and a binary value $\tilde{y}_j\in\{0,1\}$. After getting the coarse-grained binary scores $\{\tilde{y}_1,\cdots,\tilde{y}_h\}$, they are further refined by smoothing techniques (e.g., exponential moving average~\cite{yang2024follow} and Gaussian smoothing~\cite{ye2019anopcn}) to obtain a set of continuous anomaly scores for all frames as $\hat{Y} = [\hat{y}_1, \dots, \hat{y}_F]$. The key limitation in such anomaly scoring process is that the initial anomaly scoring only considers limited context and can lead to misunderstanding, which has been shown in Fig.~\ref{fig:WeaknessAddressedFigure}.

\subsection{CEA: Context-Aware Initial Anomaly Scoring}
\label{sec:CEA}

As shown in Fig.~\ref{fig:MainFigure}, \textbf{CEA} maintains a visual memory to store historical segments associated with each current segment $v_i$. By retrieving relevant memory entries, CEA constructs periodic context summaries that expand the temporal receptive field for initial anomaly scoring. The generated summaries are filtered through anomaly-aware criteria and selectively incorporated into the initial anomaly scoring process. This can be divided into four stages, as detailed as follows.

\noindent\emph{Stage 1: Dynamic Memory for Context Expansion.} In any time step $i$, to construct the context summary for the current segment $v_i$, we maintain a compact visual memory that stores representations of preceding segments $H_i=\{v_{i-n}, \dots, v_{i-1}\}$, where $n$ is the maximum number of historical segments retained in memory. 
For each previous segment $v_j\in H_i$, we extract the  $\ell_2$-normalized embedding  of the center frame $c_j$ by a pretrained image encoder $\psi(\cdot)$, denoted as $m_j = \frac{\psi(c_j)}{\|\psi(c_j)\|_2}$,  and $m_j \in \mathbb{R}^d$ where $d$ is the feature dimensionality.
Consequently, each $v_i$ has a context expansion candidate pool $M_i=\{m_{i-n},\cdots,m_{i-1}\}$.

\noindent\emph{Stage 2: Key Frame Selection by Diversity.}
Directly aggregating all embeddings in $M_i$ may introduce redundancy and dilute discriminative cues. Instead, CEA performs \emph{farthest-point sampling}~\cite{gonzalez1985clustering} in the embedding space to select a compact set of diverse key embeddings $\mathcal{S}_i \subset M_i$: Starting from an initial embedding, we iteratively select $m^* = \arg\max_{m \in M_i \setminus \mathcal{S}_i}
\min_{s \in \mathcal{S}_i} \| m - s \|_2 ,$
thereby maximizing the minimum pairwise Euclidean distance within the selected set. This strategy explicitly encourages diversity among selected embeddings. The procedure yields up to $K$ representative embeddings, providing a compact and diverse approximation of the historical context for each segment $v_i$.

\noindent\emph{Stage 3: Visual-Grounded Summary Generation.} With the given historical context, CEA will generate summary as the expanded context  periodically every $S$ segments, once at least three historical snippets are available. At each update step $j$, the selected key frames corresponding to $\mathcal{S}_j$ are fed into a VLM-based captioner $\Phi$ together with a strictly constrained prompt $\mathcal{P}$, producing a summary $s_j = \Phi(\mathcal{S}_j, \mathcal{P})$. The prompt $\mathcal{P}$ enforces four requirements: (i) describing only clearly observable visual content; (ii) avoiding speculation about intent, normality, or future events; (iii)  stating uncertainty explicitly when evidence is insufficient; and (iv) producing concise responses with 2--4 short bullet points. Importantly, $\Phi$ is explicitly prohibited from labeling events as normal or anomalous, ensuring that the generated summary serves as objective visual memory rather than semantic judgment.

\noindent\emph{Stage 4: Context-Aware Initial Scoring.} Critically, CEA is aware that inaccurate or weakly aligned summaries may degrade prediction, so it includes a \textbf{cross-modal grounding gate} to assess summary validity before incorporating it into scoring: the generated summary is used to augment anomaly inference only when it is reliably grounded in the current visual content. To specify, given a set of uniformly sampled frames $I^{(i)}=\{I_k^{(i)}\}_{k=1}^{\kappa}$ from segment $v_i$, CEA computes their $\ell_2$-normalized image embeddings $\{\mathbf{u}_k\}_{k=1}^{\kappa}$ ($\kappa$ is the number of frames) and the normalized text embedding $\mathbf{t}_i$ of the generated summary $s_i$ with CLIP~\cite{radford2021learning}. Cosine similarities are computed as $\alpha_k = \mathbf{u}_k^\top \mathbf{t}_i$ for each frame $I_k^{(i)}\in\mathbf{u}_k$ and we derive two statistics: (1) \textbf{Top-$\tilde{K}$ Mean Similarity}: $\mu_i = \frac{1}{\tilde{K}} \sum_{k \in \mathcal{T}_i} \alpha_k$, where $\mathcal{T}_i$ indexes the top-$\tilde{K}$ similarity values. This measures the strength of visual-text alignment.  (2) \textbf{Normalized Shannon Entropy}: $H_i = - \frac{1}{\log \kappa} \sum_{k=1}^{\kappa} p_k \log p_k$,  where $p_k = \mathrm{softmax}(\alpha_k / \tau)$ with temperature $\tau$. This quantifies the concentration of alignment evidence across frames.  $\mu_i$ and $H_i$ are in the range $[0,1]$.

In the gating mechanism, the summary is accepted only if $\mu_i > \delta_{\text{sim}}$ and $H_i < \delta_{\text{ent}}$, where $\delta_{\text{sim}}$ and $\delta_{\text{ent}}$ are predefined thresholds, ensuring that it is both strongly aligned with the current segment and supported by concentrated visual evidence. Validated summaries are stored as part of the memory and used as context for subsequent segments. Otherwise, the system reverts to frame-only inference without historical context. When validated, the summary is prepended to the VAD prompt and treated as a concise description of prior segments, enabling the VLM to detect deviations in $v_i$ relative to historical context. This will safely extend temporal reasoning while mitigating error accumulation from unreliable summaries. In formulation, we have the initial score $\tilde{y}_i$ and explanation $E_i$ by 

\begin{equation}
(\tilde{y}_i, E_i) =
\begin{cases}
f(I^{(i)}, s_i; \mathcal{P}_{\text{score}}^{\prime}), 
& \text{if } \mu_i > \delta_{\text{sim}} \;\land\; H_i < \delta_{\text{ent}}, \\[6pt]
f(I^{(i)}; \mathcal{P}_{\text{score}}), 
& \text{otherwise},
\end{cases}
\label{eq:initial}
\end{equation}
where $\mathcal{P}_{\text{score}}$ denotes the prompt for initial anomaly scoring without summarized context, while $\mathcal{P}_{\text{score}}^{\prime}$ augments this prompt with the validated summary $s_i$ as historical context. $E_i$ is the corresponding explanation for why $f$ assigns $\tilde{y}_i$ for $v_i$ by following the prompt instruction.

 \noindent\emph{Discussion: Computational Insight in CEA.} The introduction of memory-based CEA is necessary because existing memory paradigms are not directly suited for VAD (validated in results shown in Table~\ref{tab:ablations}): (1) Trending text-centric agentic memory systems such as A-Mem~\cite{xu2025mem} and Mem0~\cite{chhikara2025mem0} maintain and retrieve information purely in the textual embedding space. While effective for long-horizon language reasoning, they cannot perceive visual information and the abstracted historical information obscures subtle appearance-level cues that are crucial for VAD. (2) In addition, existing memory mechanisms proposed for general video understanding, such as StreamForest~\cite{zengstreamforest}, primarily focus on accumulating temporal knowledge for recognition or summarization. However, these memory systems are designed for holistic semantic accumulation and do not focus on the semantic change in the video. Contrastively, the used summary from the memory in CEA remains image-grounded, selectively updated, and explicitly validated through cross-modal alignment, ensuring high utility in VAD.

\subsection{REA: Recursive Evidence Aggregation for Anomaly Reasoning}
\label{sec:RACA}
As shown in Fig.~\ref{fig:MainFigure}, on top of CEA,  {\name} further includes a \emph{Recursive Anomaly 
Evidence Aggregation} (REA) algorithm that aggregates weak local anomaly evidence into coherent temporal intervals. Rather than treating  VAD as independent segment inference, {\name} reinterprets it as a temporal evidence aggregation problem, which can be described into 4 stages:

\noindent\emph{Stage 1: Temporal Evidence Field Construction.} Based on Eq.~\eqref{eq:initial}, CEA computes a continuous evidence-based anomaly score $y^{(ev)}_i \in [0,1]$ by considering both $\tilde{y}_i$ and  $E_i$, i.e., 
\begin{equation}
y^{(ev)}_i = \alpha \tilde{y}_i 
+ \gamma \cdot \text{Cue}(E_i) 
- \delta \cdot \text{Neg}(E_i),
\label{eq:cont}
\end{equation}
where $\tilde{y}_i\in\{0,1\}$ is the context-expanded anomaly indicator obtained from Eq.~\eqref{eq:initial}, $\text{Cue}(E_i)\in\mathbb{Z}_{\ge 0}$ counts the number of matches to predefined anomaly-cue keywords in the explanation $E_i$, and $\text{Neg}(E_i)\in\mathbb{Z}_{\ge 0}$ counts the number of matches to predefined anomaly-denial keywords in $E_i$. The keyword sets are dataset-agnostic and fixed across all datasets.
$\alpha,\gamma,\text{and} \ \delta$ are weighting coefficients (hyperparameters). Eq.~\eqref{eq:cont} will produce a continuous anomaly evidence landscape over time that integrates two complementary sources: (i) the implicit inference results output from $f$ indicated by $\tilde{y}_i$, and (ii) the explicit causal inference expressed by $f$ indicated by $E_i$ either supporting or contradicting evidence. Such score assignment will express $f$'s belief on the existence of an anomaly more faithfully. 

\noindent\emph{Stage 2: Anomalous Window Proposals.}
Given the segment-level initial scores $[y^{(ev)}_1,\cdots,y^{(ev)}_h]$ by iterating Eq.~\eqref{eq:cont} from $i=1$ to $h$, REA constructs temporal window proposals via recursive binary partitioning. To specify, starting from the full interval $W^{(0)} = [0, h]$, REA iteratively splits each window $W = [i_{\text{start}}, i_{\text{end}}]$ at its midpoint  $i_{\text{mid}} = \left\lfloor \frac{i_{\text{start}} + i_{\text{end}}}{2} \right\rfloor$,
yielding two sub-windows $[i_{\text{start}}, i_{\text{mid}}]$ and $[i_{\text{mid}}+1, i_{\text{end}}]$. Specifically, in the general case, given any proposed window $W=[i_{start},i_{end}]$ ($0\le i_{start}< i_{end}\le h$) covering from the $i_{start}$-th segment to the $i_{end}$-th segment, REA computes statistics to represent the anomaly potential: 

\begin{equation}
\bar{y}_W = \frac{1}{|W|}\sum_{i \in W} y^{(ev)}_i,
\quad
y^{\max}_W = \max_{i \in W} y^{(ev)}_i,
\label{eq:stats}
\end{equation}
 
In practice, a window is considered anomalous if it exhibits either a sufficiently strong local peak ($y_W^{\max}$) or sustained anomaly ($\bar{y}_W$). High $y^{\max}_W$ indicates the presence of a sharp anomalous peak within $W$, while $\bar{y}_W$ reflects the persistence or density of anomalous evidence across segments.

This recursive binary partitioning proceeds only along windows that satisfy the anomaly criteria in Eq.~\eqref{eq:stats}. Recursion along a branch terminates when either
(1) the window length $|W|$ falls below a predefined minimum size $l_{min}$,
(2) a maximum recursion depth $d_{max}$ is reached, or
(3) the window no longer satisfies the anomaly criteria.
Such hierarchical splitting produces a multi-scale set of candidate temporal intervals corresponding to clusters of anomaly evidence. Meanwhile, adjacent or overlapping candidate intervals generated by recursion are subsequently merged to enforce temporal continuity before ranking. Please see Appendix~\ref{app:pseudocodes} for pseudocodes and more details.

\noindent\emph{Stage 3: Dynamic Proposal Selection.} If the initial window $W^{(0)}$ fails the anomaly criteria defined by $\bar{y}_W$ and $y_W^{\max}$, recursion stops and no further subdivision is performed. Otherwise, when it is terminated, candidate windows are ranked according to their cumulative anomaly evidence:  
\begin{equation}
S(l,r) = \sum_{i=l}^{r} y^{(ev)}_i,
\label{eq:evidence}
\end{equation}

Candidate windows are ranked according to their cumulative anomaly evidence in Eq.~\eqref{eq:evidence}, and the Top-$K_{\max}$ highest-scoring intervals are retained. The final anomalous regions selected for further investigation are
\begin{equation}
    \mathcal{C} = \{C_m = [l_m, r_m]\}_{m=1}^{K_{\max}},
    \label{eq:candidate}
\end{equation}
which contains a set of $K_{\max}$ anomalous windows for video $V$.

\noindent\emph{Stage 4: Event-Based Explanation-Augmented VAD.}
\label{sec:cluster_explain}
Based on segment-level initial scores $[y^{(ev)}_1,\cdots,y^{(ev)}_h]$, {\name} gets a new set of frame-level scores $\hat{Y}^{(ev)} = [\hat{y}_1^{(ev)}, \dots, \hat{y}_F^{(ev)}]$ by following the coarse-to-fine score post-processing in VERA~\cite{ye2024vera} as the desired output for VAD. 
The difference lies in that {\name} will output a new set of event-based explanations for the test video $V$  given  the set $\mathcal{C}$ output from  Eq.~\eqref{eq:candidate}.  
Specifically, let $C_m=[l_m,r_m]$ denote an anomalous interval discovered by REA. Rather than generating explanations independently per segment like existing pipelines, {\name} treats the frames spanning the cluster $C_m$ as a coherent event and derive the text description from previous representative segment explanations. Recall that for each segment $v_i$ we store a single textual explanation $E_i$ produced by $f$ during inference in Eq.~\eqref{eq:initial}. To ground the generated event-level explanation in the internal evidence pattern of $C_m$,
REA selects representative segments $\mathcal{S}_m \subseteq [l_m,r_m]$ with all segments satisfying any of the following criteria to capture its onset, peak intensity, and temporal extent (with maximum 10 segments): (1) the segment locates in the temporal boundaries, (2) the segment contains high evidence scores $y^{(ev)}_i$, or (3) the segment is from transition segments that mark onset and resolution.  

After that, REA will generate \textbf{visual–textual grounded event-level explanation}. To specify, for an anomaly cluster $C_m$, we sample $\kappa$ representative frames uniformly across its temporal span (denoted $\{I^{(i)}_m\}_{i=1}^{\kappa}$) and pair them with the textual evidence $\{\, E_i \mid i \in \mathcal{S}_m \,\}$. These multi-modal inputs well represent the event, and an event-level explanation is then generated by:
\begin{equation}
\hat{E}_m = f(\{I^{(i)}_m\}_{i=1}^{\kappa}, \{E_i|i \in \mathcal{S}_m\};\mathcal{P}_{caption}),
\label{eq:caption}
\end{equation}
where $\mathcal{P}_{caption}$ is a prompt instructing $f$ to produce a concise ($\leq4$ sentence) narrative strictly grounded in the provided evidence.
By enforcing both visual anchoring and evidence selection, the generated explanation remains aligned with the underlying anomaly cluster.  

\noindent\emph{Discussion: Computational Insight in REA.} When we iterate Eq.~\eqref{eq:caption} for all identified anomalous windows, {\name} will output the desired event-based explanation and accurate anomaly scores, as validated by quantitative and qualitative results in \S~\ref{sec:exp_result}. Importantly, under this formulation, localization and explanation are not disjoint stages but two complementary operations over the same temporal anomaly evidence field. That is, explanations from {\name} are derived directly from the temporal structure uncovered by recursive evidence aggregation based on temporal anomaly field $\{y^{(ev)}_i\}_{i=1}^{h}$. This unified design promotes temporal coherence, reduces redundancy inherent in segment-level rationales, and ensures that each explanation is explicitly grounded in the anomaly evidence that triggered its selection. 
\section{Experiments and Results}
\label{sec:exp_result}

We evaluate three aspects of {\name}: (Q1) \textbf{Accurate Detection}: Does the proposed  {\name}  improve VAD performance? (Q2) \textbf{Faithful Explainability}: Does {\name} provide event-based explanations that are more consistent and informative than existing VAD reasoning?  And (Q3) \textbf{Modular Effectiveness}: Do CEA and REA in {\name} produce reliable context-aware anomaly scores and convert fragmented predictions into temporally coherent anomaly intervals?

\subsection{Experimental Settings}
\label{sec:exp}
\noindent\emph{Datasets.}
We evaluate model performance on two large-scale benchmarks: UCF-Crime \cite{sultani2018real} and XD-Violence \cite{wu2020not} following recent studies~\cite{ye2024vera,zanella2024harnessing}. UCF-Crime contains 13 anomaly types and a test set of 290 videos (140 abnormal), with an average duration of 2.13 mins. XD-Violence contains 6 anomaly categories and a test set of 800 videos (500 abnormal), with an average duration of 1.62 mins.

\noindent\emph{Baselines.}
We adopt representative explainable VAD methods including Holmes-VAD~\cite{zhang2024holmes}, LAVAD~\cite{zanella2024harnessing}, LLAVA~\cite{liu2024visual}, VADor~\cite{lv2024video}, and VERA~\cite{ye2024vera} as baselines. For explainability, we compare our event-level explanations to three baseline segment-level explanations from VERA within the predicted anomaly event: the most anomalous segment according to Eq. \eqref{eq:cont}, a random segment explanation, and concatenated segment-level explanations. \textbf{Please note that since previous methods~\cite{ye2024vera,zanella2024harnessing} have proved the effectiveness of using frozen VLMs without instruction tuning or fine-tuning in VAD, the performance is reported with frozen VLMs}.

\noindent\emph{Evaluation metrics.}
For detection accuracy, we report frame-level  Area Under the ROC Curve (AUC) on both datasets and Average Precision (AP) on XD-Violence following standard VAD practice.
For explainability, since those benchmarks do not provide ground truth textual rationales, we evaluate generated explanations (with 200 random examples drawn from UCF-Crime) with three complementary proxies:
(i) \textbf{Human preference} (\%), where \underline{human raters} choose which explanation better describes the anomalous event while staying faithful to visible evidence;
(ii) \textbf{Conciseness} measured by average token length; and
(iii) \textbf{Descriptiveness} measured by anomaly-category prediction accuracy (\%), where an \underline{external LLM} predicts the anomaly class given \emph{only} the generated explanation text without frames. This measures whether explanations contain discriminative event details rather than generic anomaly statements.

\noindent\emph{Implementation details.}
By default, we use InternVL2-8B as the base VLM $f$ for {\name} as \cite{ye2024vera} does. Videos are partitioned into temporal segments every 16 frames as \cite{ye2024vera, zanella2024harnessing} do, each represented by $\kappa=8$ uniformly sampled frames.
For CEA, the image encoder $\psi(\cdot)$ is ResNet-50 to extract the embeddings. We retain at most $n=8$ historical segment embeddings in memory and select $K=4$ diverse key frames.
Summaries are refreshed every $S=5$ segments and generated by a VLM captioner $\Phi$ (InternVL2-8B).
For grounding, each current segment uses $\kappa=8$ sampled frames to compute image-text alignment with the summary with CLIP embeddings; we adopt the gating thresholds $(\delta_{\text{sim}},\delta_{\text{ent}})=(0.30,0.80)$ unless otherwise stated. For REA, it retains at most $k_{\max}=6$ proposals.

\begin{table}[tb]
  \caption{Comparison with leading explainable VAD methods. Performance is reported using frozen VLMs without any instruction tuning or fine-tuning because frozen VLMs have been shown to be effective for VAD tasks by~\cite{ye2024vera,zanella2024harnessing}. $^{\dagger}$ Note that Holmes-VAD requires limited task-specific training, and is therefore not strictly frozen.}
  \label{tab:results}
  \centering
  \begin{tabular}{lccc}
    \toprule
    & {AUC (\%)} & {AUC (\%)} & AP (\%)\\
    \cmidrule(lr){2-4}
    Method & UCF-Crime &  \multicolumn{2}{c}{XD-Violence} \\
    \midrule
    Holmes-VAD \cite{zhang2024holmes}$^{\dagger}$ & 84.61 & Not Reported & \textbf{84.96} \\
    LAVAD \cite{zanella2024harnessing} & 80.28 & 85.36 &  62.01\\
    LLAVA-1.5 \cite{liu2024improved} & 72.84 & 79.62 & 50.26\\
    VADor \cite{lv2024video} & 85.90 &  Not Reported &  Not Reported \\
    VERA \cite{ye2024vera} & 86.55 & 88.26 & 70.54\\
    {\name} (Our Approach) & \textbf{87.63} & \textbf{90.14} & 74.68\\
    \bottomrule
  \end{tabular}
\end{table}
\subsection{Comparison in Accuracy and Explainability}
\label{sec:comparison}
We first answer (Q1) by comparing {\name} with state-of-the-art explainable VAD methods. The results are shown in Table~\ref{tab:results}. The frame-level AUC on both datasets on Table~\ref{tab:results} shows that {\name}      achieves the best AUC performance on UCF-Crime and XD-Violence, improving over the strongest baseline VERA \cite{ye2024vera} by +1.08\% and +1.88\% AUC points respectively.  In addition, {\name} obtains a 4.14\% AP improvement over VERA. Compared to VERA, {\name} employs the context-expanded and evidence-grounded scoring formulation in Eq.~\eqref{eq:initial}.  These results support our central hypothesis that anomalies are best detected as deviations from an evolving temporal narrative and demonstrates the effectiveness of incorporating structured temporal context into anomaly inference.

We next answer (Q2) on whether event-level explanations by {\name} are of higher quality. Table~\ref{tab:explanations} compares our event-level explanations with several segment-level alternatives by VERA: the explanation from the single most anomalous segment, a random anomalous segment, and concatenation of all segment explanations within the detected event by {\name}.
Results show that event-level explanations from {\name} are strongly preferred by humans (80.5\%) while being substantially shorter (46.6 tokens on average), suggesting they better capture the \emph{complete event}  than a narrow or noisy snapshot. They also achieve the highest category-prediction accuracy (27.91\%), indicating that aggregating evidence across the anomalous window yields more discriminative descriptions than isolated segments. As for the concatenation baseline, although it achieves relatively high prediction accuracy, the explanation is prohibitively long (2409.5 tokens), which highlights the necessity of introducing structured temporal aggregation for explanation semantics as {\name} does.

\begin{table}[tb]
  \caption{We compare the  event-level explanations generated by {\name}  with previous segment-level explanations generated by VERA on human evaluation preference percentage, anomaly category prediction accuracy, and average token length.}
  \label{tab:explanations}
  \centering
  \begin{tabular}{@{}lccc@{}}
    \toprule
    Explanation Type & Human Eval. (\%) & Pred. Acc. (\%) & Token Length\\
    \midrule
    Event-Level & \textbf{80.5} & \textbf{27.91} & \textbf{46.6} \\
    Most Anomalous Segment-Level & 11.5 & 24.62 & 68.9 \\
    Random Segment-Level & 6.0 & 16.26 & 71.2 \\
    All Segment-Level Concatenated & 2.0 & 26.37 & 2409.5  \\
  \bottomrule
  \end{tabular}
\end{table}

\begin{table}[tb]
\centering
\caption{Experiment results verifying methodology design in {\name} on UCF-Crime. Left: Ablation study results on CEA. Right: Ablation study results on REA.}
\label{tab:ablations}

\small
\setlength{\tabcolsep}{4pt}

\newlength{\ablationblockheight}
\setlength{\ablationblockheight}{5.2cm}

\begin{minipage}[t][\ablationblockheight][t]{0.48\linewidth}
\centering
\begin{tabular}{lc}
\toprule
\multicolumn{2}{c}{\textbf{CEA Ablations}} \\
\midrule
Method & AUC (\%) \\
\midrule
InternVL2-8B (Base) & 86.42 \\
+ Non-gated Summaries & 84.72  \\
+ KLE \cite{nikitin2024kernel} Gated & 86.28 \\
+ Shannon Entropy Gated & 86.61 \\
+ Similarity Gated & 85.86 \\
+ Similarity \& Entropy\ Gated & \textbf{86.68} \\
\midrule
A-Mem-Based Frame Retrieval & 84.07 \\
StreamForest  Compression & 78.73  \\
\bottomrule
\end{tabular}
\vfill 
\end{minipage}
\hfill
\begin{minipage}[t][\ablationblockheight][t]{0.48\linewidth}
\centering
\begin{tabular}{lc}
\toprule
\multicolumn{2}{c}{\textbf{REA Ablations}} \\
\midrule
Method & AUC (\%) \\
\midrule
REA & \textbf{87.63}  \\
-- w/o $\alpha \tilde{y}_i $ & 67.02 \\ 
-- w/o $\gamma \cdot \text{Cue}(E_i)$ & 87.60 \\ 
-- w/o $\delta \cdot \text{Neg}(E_i)$ & 87.32 \\ 
-- w/o $\gamma \text{Cue}(E_i) - \delta\text{Neg}(E_i)$  & 87.36 \\ 
-- w/o Dynamic Selection & 86.77 \\
\midrule
Non-Gated Summaries + REA & 85.26 \\
No Summaries + REA & 87.14 \\
\bottomrule
\end{tabular}
\vfill
\end{minipage}

\end{table}

\subsection{Ablation Studies for Methodology Design}
Without loss of generality, we perform ablation studies with InternVL2-8B as $f$ on UCF-Crime to answer (Q3), with results shown in Table~\ref{tab:ablations}. We derive the following findings:

\noindent (1) \emph{Gating in CEA is essential for reliable context expansion.} Directly incorporating historical summaries without validation degrades performance (Non-gated Summaries: 84.72\% vs.\ Base: 86.42\% in AUC), indicating that unfiltered context may introduce error accumulation from misaligned or weakly grounded memory. In contrast, the proposed similarity- and entropy-based grounding gate restores and further improves performance (86.68\%), demonstrating that contextual expansion is beneficial only when supported by strong visual–text alignment. 

\noindent (2) \emph{More complex language-level uncertainty measures do not necessarily translate to improved cross-modal grounding in CEA.} We further investigate different grounding strategies for context validation. Kernel Language Entropy (KLE)~\cite{nikitin2024kernel}, which is an expansion of semantic entropy \cite{farquhar2024detecting, kossen2024semantic, kuhn2023semantic} used for uncertainty estimation in language models, achieves 86.28\% AUC, which is inferior to both Shannon entropy gating (86.61\%) and the proposed similarity–entropy joint gating (86.68\%). Thus, directly modeling visual–text alignment concentration via entropy over similarity scores provides a more effective signal for filtering unreliable summaries.

\noindent \emph{(3) Image-grounded and deviation-aware mechanisms are important for memory design in CEA.} We additionally compare against text-centric memory mechanisms (A-Mem retrieval~\cite{xu2025mem}) and memory frameworks designed for general video understanding (StreamForest~\cite{zengstreamforest}). Despite their increased computational overhead, these methods underperform our approach, demonstrating that effective memory design for VAD must be both image-grounded and explicitly deviation-aware as the one proposed in CEA.

\noindent\emph{(4) Each component in REA contributes to robust temporal aggregation.}
The full use of REA achieves the best performance (87.63\% AUC). Removing any core component consistently degrades performance. In particular, eliminating the anomalous score weighting ($\tilde{y}_i$) leads to a dramatic drop to 67.02\%, indicating that segment-level anomaly evidence is fundamental to effective aggregation. Removing the cue-word evidence term $\text{Cue}(E_i)$ results in a slight decrease to 87.60\%, while removing the negative evidence term $\text{Neg}(E_i)$ lowers performance to 87.32\%. Disabling dynamic proposal selection causes a more noticeable reduction to 86.77\%, highlighting its importance in suppressing spurious intervals and enforcing temporal coherence. These results confirm that each component plays a complementary role in structured evidence aggregation.

Furthermore, REA remains effective even without using summaries (87.14\%), outperforming gated summaries alone (86.68\%), demonstrating that recursive temporal aggregation itself substantially improves localization stability. However, combining REA with non-gated summaries significantly harms performance (85.26\%), reinforcing that reliable context validation is essential for {\name}.

\begin{table}[tb]
  \caption{Average number of predicted anomalous events per video on UCF-Crime.
  }
  \label{tab:chunks}
  \centering
  \begin{tabular}{@{}lcc@{}}
    \toprule
    Method & w/ Summaries from CEA & w/o Summaries from CEA\\
    \midrule
    With Aggregation & 2.53 & 2.19\\
    Without Aggregation & 14.92 & 13.91 \\
  \bottomrule
  \end{tabular}
\end{table}

\noindent\emph{(5) Event-level aggregation in REA reduces temporal fragmentation.} Segment-wise VAD predictions produce many short, fragmented anomaly intervals  across adjacent video segment.
Contrastingly, {\name} overcomes this dilemma with REA: Additional results shown in Table~\ref{tab:chunks} demonstrate  that REA dramatically reduces the number of predicted anomalous intervals per video (e.g., from 14.92 to 2.53 when summaries are used), indicating that REA suppresses transient spikes and consolidates evidence into coherent event windows. Importantly, this is achieved without sacrificing detection accuracy and can help improve the explanation quality, as \S~\ref{sec:comparison} have shown.

\begin{figure}[tb]
    \centering
    
    \begin{subfigure}[t]{0.45\linewidth}
        \centering
        \includegraphics[width=\linewidth]{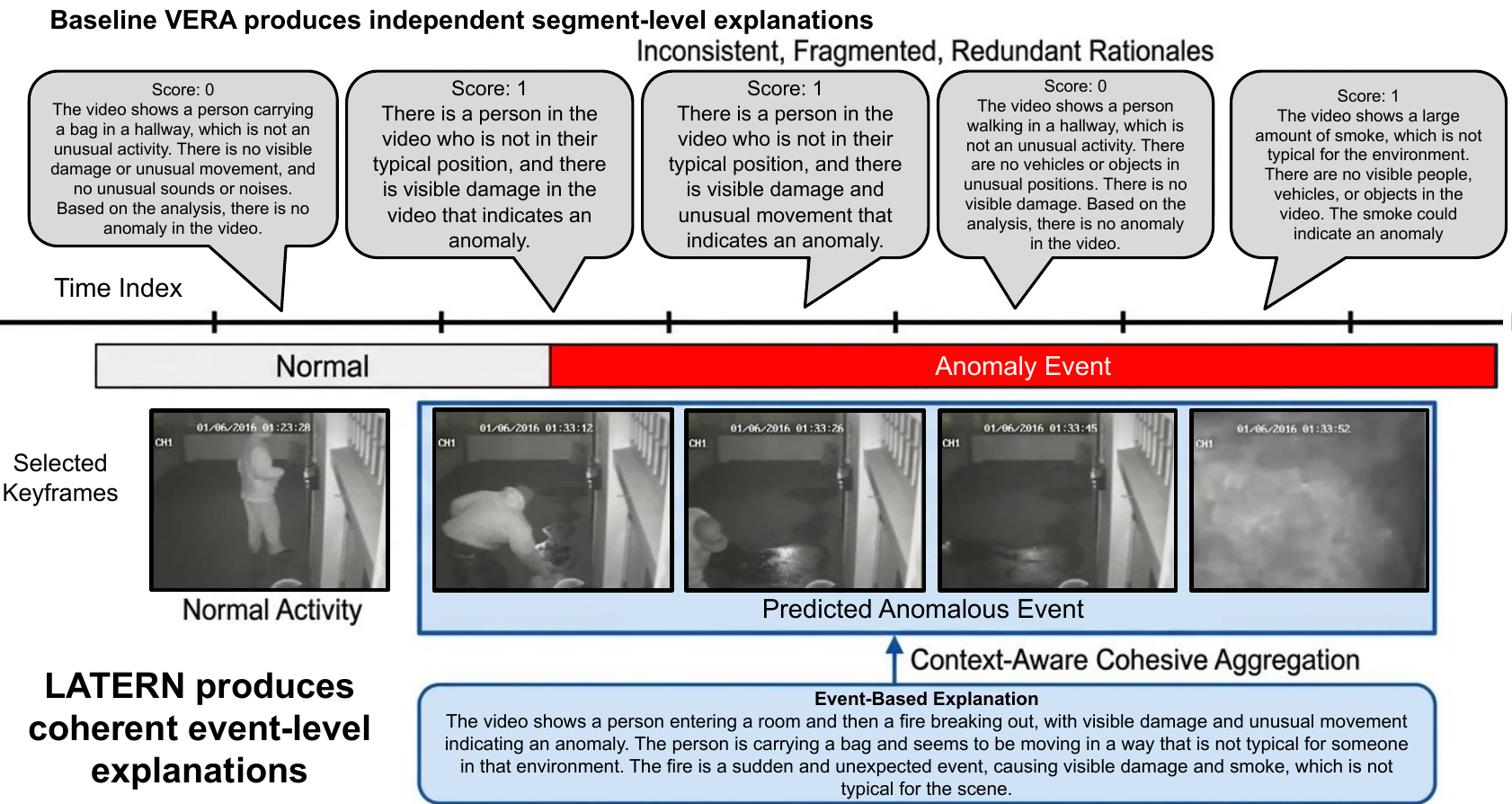}
        \caption{Qualitative Example of video ``Arson041\_x264'' in UCF-Crime comparing generated segment-level explanations to event-level explanations by {\name}.}
        \label{fig:EventBasedExplanation}
    \end{subfigure}
    \hfill
    \begin{subfigure}[t]{0.54\linewidth}
        \centering
        \includegraphics[width=\linewidth]{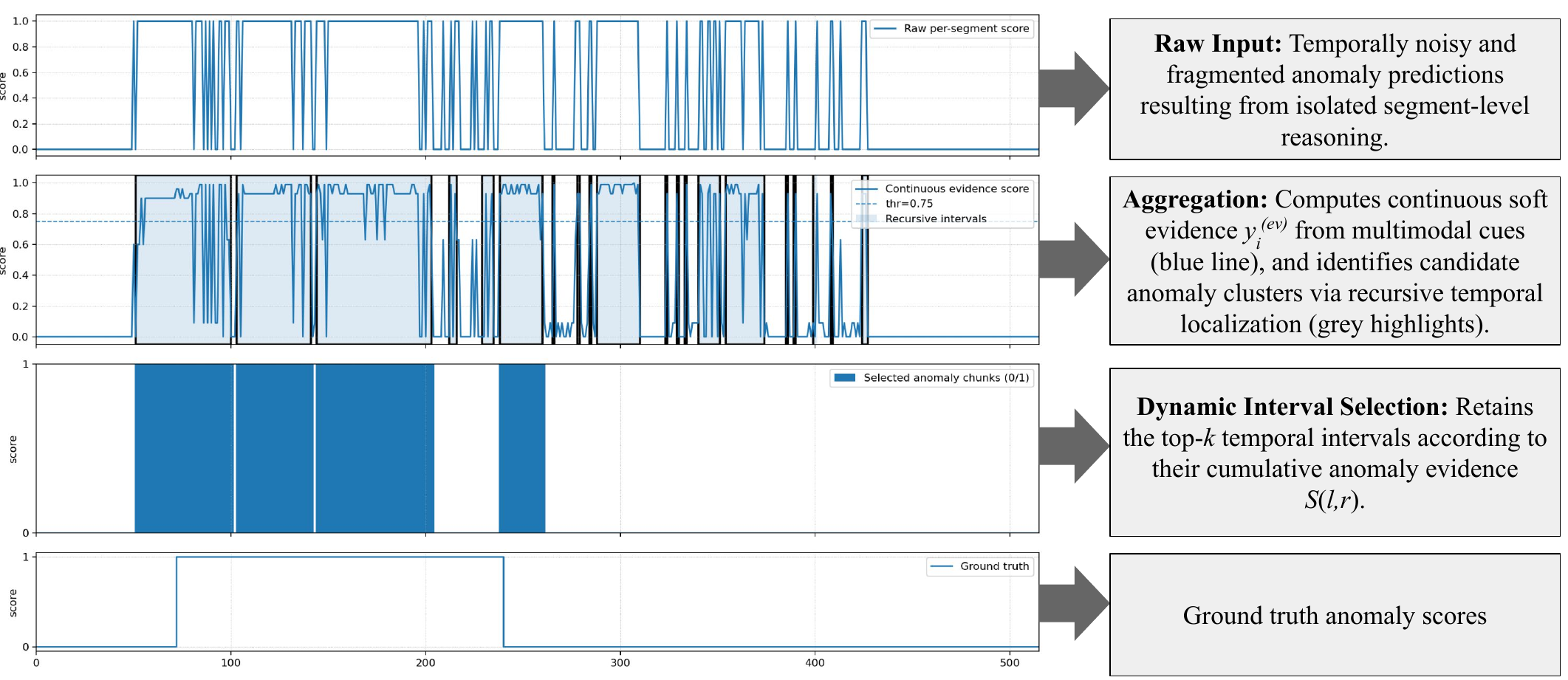}
        \caption{REA can generate anomalous event intervals given initial segment-level anomaly scores which better aligns with ground truth anomaly scores.}
        \label{fig:RACA_example}
    \end{subfigure}

    \caption{(1) Comparison between segment-level and event-level explanations produced by {\name}. (2) REA converts fluctuating segment scores into coherent anomaly intervals aligned with ground truth.}
    \label{fig:combined_example}
\end{figure}

\subsection{Qualitative Results and Case Studies}
We would like to provide qualitative results as follows to demonstrate the effectiveness of {\name} in generating event-based explanations. First, Fig.~\ref{fig:EventBasedExplanation} illustrates that existing methods (exemplified by VERA) output segment-level explanations which fluctuate across neighboring segments and often provide partial, inconsistent descriptions of the same underlying event. In contrast, {\name} produces a single event-level explanation aligned with the temporally consolidated anomaly interval, summarizing onset, evolution, and the key visible evidence. This qualitative behavior mirrors our quantitative results: {\name} improves temporal coherence, and cluster-level semantic aggregation produces explanations that are shorter yet more informative than segment-level rationales.

Additionally, Fig.~\ref{fig:RACA_example} takes an exemplary video (``Burglary032\_x264'' in UCF-Crime) to illustrate how REA transforms unstable segment-level predictions into coherent event-level anomaly localization. The initial context-aware scores exhibit noticeable fluctuations across adjacent segments, producing fragmented high-response regions. Through recursive evidence aggregation, REA generates multiple candidate anomaly windows at different temporal scales, progressively organizing dispersed anomaly signals into structured clusters.  After dynamic interval selection, spurious short intervals are suppressed, and the remaining window forms a temporally coherent anomaly region that closely aligns with the ground-truth event boundaries.

\section{Concluding Remarks}
Overall, {\name} demonstrates that effective video anomaly detection with VLMs requires moving beyond isolated segment-level reasoning toward structured temporal evidence aggregation. By integrating image-grounded memory with recursive interval-level reasoning, {\name}  bridges detection and explanation under a unified context-aware formulation. Extensive experimental results on representative benchmarks demonstrate that incorporating structured temporal aggregation substantially enhances frozen VLMs, highlighting the potential of temporally coherent and context-aware reasoning for complex video understanding.

\clearpage
\appendix

\begin{center}
	{\Large \textbf{Supplementary Material for}}\\[0.5em]
	{\large \textbf{{\name}: Test-Time Context-Aware Explainable Video Anomaly Detection}}
\end{center}

\section{Pseudocodes}
\label{app:pseudocodes}

\renewcommand{\thefigure}{A.\arabic{figure}}
\renewcommand{\thetable}{A.\arabic{table}}
\renewcommand{\theequation}{A.\arabic{equation}}

\setcounter{figure}{0}
\setcounter{table}{0}
\setcounter{equation}{0}

\begin{algorithm}[t]
	\caption{Context-Aware Anomaly Scoring (CEA)}
	\label{alg:cea}
	\begin{algorithmic}[1]
		\Require Video $V$, segment index set $\{t_i\}_{i=1}^{h}$, summary stride $S$, minimum history length $m_{\min}$, history capacity $M$
		\Ensure Segment-level predictions $\{(\hat{y}_i, r_i, u_i)\}_{i=1}^{h}$
		
		\State Initialize history buffer $\mathcal{H} \gets \emptyset$
		\State Initialize history summary $u \gets$ ``No prior events observed yet''
		\State Initialize snippet counter $c \gets 0$
		
		\For{$i=1$ to $h$}
		\State Extract current temporal segment $x_i$ from $V$ centered at $t_i$
		\State Sample frames from $x_i$ and construct segment visual input
		\State Select representative frame $\tilde{x}_i$ from $x_i$
		\State Update history buffer: $\mathcal{H} \gets \textsc{UpdateHistory}(\mathcal{H}, \tilde{x}_i, t_i, M)$
		\State $c \gets c + 1$
		
		\State $\texttt{use\_summary} \gets \textbf{False}$
		\If{$|\mathcal{H}| \ge m_{\min}$ \textbf{and} $(c \bmod S)=0$}
		\State $u \gets \textsc{SummarizeHistory}(\mathcal{H})$
		\State $\texttt{use\_summary} \gets \textsc{GroundAndGate}(u,\mathcal{H})$
		\EndIf
		
		\If{$\texttt{use\_summary}$}
		\State $(\hat{y}_i, r_i) \gets \textsc{ScoreWithSummary}(x_i, u)$
		\Else
		\State $(\hat{y}_i, r_i) \gets \textsc{ScoreOriginal}(x_i)$
		\EndIf
		\EndFor
		
		\State \Return $\{(\hat{y}_i, r_i, u_i)\}_{i=1}^{h}$
	\end{algorithmic}
\end{algorithm}
\begin{algorithm}[t]
	\caption{Recursive Evidence Aggregation (REA)}
	\label{alg:rea}
	\begin{algorithmic}[1]
		\Require Segment predictions $\mathcal{S}=\{s_i\}_{i=1}^{h}$, maximum intervals $K_{\max}$, minimum window length $l_{\min}$, maximum recursion depth $d_{\max}$, merge gap $\delta$
		\Ensure Final anomaly scores $\{\hat{y}_i\}_{i=1}^{h}$
		
		\State $\mathcal{I} \gets \textsc{RecurseLocalize}(\mathcal{S},1,h,l_{\min},0,d_{\max})$
		\State $\mathcal{I} \gets \textsc{MergeIntervals}(\mathcal{I},\delta)$
		
		\State $\mathcal{C} \gets \textsc{SelectTopKIntervals}(\mathcal{S},\mathcal{I},K_{\max})$
		
		\For{$i=1$ to $h$}
		\State $\hat{y}_i \gets \textsc{LocalEvidenceScore}(s_i)$
		\EndFor
		
		\State \Return $\{\hat{y}_i\}_{i=1}^{h}$
	\end{algorithmic}
\end{algorithm}

\begin{algorithm}[t]
	\caption{Recursive anomalous window proposal generation}
	\label{alg:recursive}
	\begin{algorithmic}[1]
		
		\Function{RecurseLocalize}{$\mathcal{S}, i_0, i_1, l_{\min}, d, d_{\max}$}
		
		\If{$i_0 > i_1$}
		\State \Return $\emptyset$
		\EndIf
		
		\State $\mathbf{z} \gets \textsc{WindowEvidence}(\mathcal{S},i_0,i_1)$
		
		\If{$\neg \textsc{LikelyAnomalousWindow}(\mathbf{z})$}
		\State \Return $\emptyset$
		\EndIf
		
		\If{$d \ge d_{\max}$ \textbf{or} $(i_1-i_0+1)\le l_{\min}$}
		\State \Return $\{(i_0,i_1)\}$
		\EndIf
		
		\State $i_m \gets \lfloor (i_0+i_1)/2 \rfloor$
		
		\State $\mathcal{I}_L \gets \textsc{RecurseLocalize}(\mathcal{S},i_0,i_m,l_{\min},d+1,d_{\max})$
		\State $\mathcal{I}_R \gets \textsc{RecurseLocalize}(\mathcal{S},i_m+1,i_1,l_{\min},d+1,d_{\max})$
		
		\State \Return $\textsc{MergeIntervals}(\mathcal{I}_L \cup \mathcal{I}_R,1)$
		
		\EndFunction
	\end{algorithmic}
\end{algorithm}

\begin{figure}[t]
	\centering
	\footnotesize
	\begin{tcolorbox}[colback=gray!5,colframe=black,boxrule=0.5pt,width=\linewidth]
		\textbf{Anomalous Cue Words and Negation Patterns used in REA $y^{(ev)}_i$}
		
		\vspace{5pt}
		\scriptsize
		\begin{minipage}[t]{0.48\linewidth}
			\textbf{Cue Keywords ($\text{Cue}(E_i)$)}\\
			\noindent fight, fighting, assault, attack, hit, punch, kick, stab, shoot, gun, weapon,
			rob, robbery, steal, stealing, theft, burglary, break in, breaking, vandal,
			vandalism, arson, fire, explosion, explode, crash, collision, accident,
			chase, chasing, running, panic, scream, blood, knife,
			climbing over a fence, climb over a fence, trespass, trespassing.
		\end{minipage}
		\hfill
		\begin{minipage}[t]{0.48\linewidth}
			\textbf{Negation Patterns ($\text{Neg}(E_i)$)}\\
			\texttt{\textbackslash b no anomaly\textbackslash b}\\
			\texttt{\textbackslash b there is no anomaly\textbackslash b}\\
			\texttt{\textbackslash b no unusual\textbackslash b}\\
			\texttt{\textbackslash b no (visible )?damage\textbackslash b}\\
			\texttt{\textbackslash b no (unusual|abnormal) (movement|events)\textbackslash b}
		\end{minipage}
		
		\vspace{5pt}
		
		\scriptsize
		Cue keywords activate semantic anomaly evidence (weighted by $\gamma$),
		while negation patterns suppress false positives (weighted by $\delta$),
		as defined in Eq.~(2).
	\end{tcolorbox}
	\caption{Fixed anomaly cue words and negation patterns used to compute $y^{(ev)}_i$}
	\label{fig:raca_keywords_negations}
\end{figure}
This section provides pseudocode for the proposed Context-Aware Anomaly Scoring (CEA) and Recursive Evidence Aggregation (REA) procedures. 
Given a sequence of temporal segments $\mathcal{S}=\{s_i\}_{i=1}^{h}$, CEA first performs context-aware segment-level anomaly analysis by maintaining a bounded history of prior segments, periodically generating a history summary, and using the summary only when it is sufficiently grounded in the visual history. The resulting segment-level predictions and corresponding VLM responses are then refined by REA, which recursively identifies candidate anomalous intervals and aggregates evidence across neighboring segments.

Algorithm~\ref{alg:cea} summarizes the overall CEA pipeline. For each segment, the method updates the history buffer with a representative frame from the current segment. When sufficient history has been accumulated and the summary refresh stride is met, CEA generates a concise summary of prior events and uses it for anomaly scoring only if the summary passes the grounding-based gating criterion. Otherwise, the method falls back to the original segment-level inference without contextual summary input.

Algorithm~\ref{alg:rea} summarizes the overall REA pipeline. Starting from the full temporal window, the algorithm recursively partitions the interval using \textsc{RecurseLocalize} (detailed in Algorithm~\ref{alg:recursive}) and retains windows that satisfy the anomaly evidence criteria. Each candidate window is evaluated using summary statistics of the local evidence scores within the window. The recursion stops once a minimum window length or maximum depth is reached, producing a set of candidate anomalous intervals.

The local evidence score for each segment combines the segment-level anomaly prediction with textual signals extracted from the VLM responses, including anomaly cue keyword matches and negation patterns. The resulting score is clipped to the range $[0,1]$.

After recursion, candidate intervals are merged to remove small gaps and up to $K_{\max}$ intervals are retained based on their accumulated evidence. The final refined anomaly scores $\{\hat{y}_i\}_{i=1}^{h}$ are then obtained by assigning each segment its corresponding local evidence score. The threshold values used in the window evidence criteria are fixed across datasets.

\subsection{REA Anomaly Cue Words and Negation Patterns}
\label{app:cue_words}

To incorporate semantic signals from the VLM-generated explanations, REA utilizes predefined anomaly cue keywords and negation patterns when computing the evidence score $y_i^{(ev)}$ described in Eq.~(2). The cue keywords correspond to terms commonly associated with anomalous activities (e.g., violence, accidents, or suspicious behavior), while the negation patterns capture explicit statements indicating the absence of anomalies.

For a given segment explanation $E_i$, the function $\text{Cue}(E_i)$ counts the number of matches to anomaly cue keywords appearing in the explanation text. Conversely, $\text{Neg}(E_i)$ counts the number of matches to predefined negation patterns that indicate normal behavior or explicitly deny the presence of anomalies. These two signals provide complementary semantic evidence that helps refine the anomaly score produced by the VLM.

The keyword sets are designed to be general and dataset-agnostic, and remain fixed across all datasets used in our experiments. This avoids dataset-specific prompt engineering and ensures consistent evidence aggregation during inference.

Figure~\ref{fig:raca_keywords_negations} lists the anomaly cue keywords and negation patterns used in REA.

\section{Additional Results and Analysis}
\label{app:more_results}

This section reports additional ablations and hyperparameter analysis that informed the final design choices. Unless stated otherwise, the numbers below are reported on UCF-Crime using the same backbone (InternVL2-8B) and inference protocol as the main paper.

\subsection{CEA Threshold/Hyperparameter Analysis}
We perform a threshold analysis on the thresholds for the thresholds introduced in the summary gating mechanism. Table \ref{tab:gate_grid_auc} shows how the mean similarity threshold and the entropy threshold effect the calculated respective AUC values \emph{before} REA has been applied. Overall, we notice that performance is stable across a large range of thresholds, indicating that gate tuning is not sensitive with respect to threshold setting.

We further study the effect of the summary stride parameter $S$ used in CEA. 
As shown in Table~\ref{tab:summaryStride}, the method is robust to the choice of stride length, with only minor performance variations across different settings. 
Interestingly, both smaller and larger stride values can yield slight improvements, suggesting that the contextual summaries provide useful information even when updated less frequently. 
At the same time, larger stride values reduce the frequency of summary updates and therefore decrease the additional computation introduced by CEA while maintaining comparable performance. 
We use $S=5$ in our experiments as a balanced default across datasets and models.

\begin{table}[t]
	\centering
	\setlength{\tabcolsep}{5pt}
	\renewcommand{\arraystretch}{1.15}
	\caption{\textbf{Gate threshold sensitivity (AUC) and Summary Stride Length.} Mean similarity threshold is on the x-axis and normalized entropy threshold on the y-axis. Performance is stable across a broad range (mean $\in[0.10,0.35]$, entropy $\in[0.60,0.85]$), indicating that gate tuning is not critical. We use \textbf{(0.30, 0.80)} in all experiments.}
	\small
	\begin{tabular}{c|cccccc}
		\toprule
		\textbf{Entropy} $\downarrow$ \textbf{/ Mean} $\rightarrow$ 
		& 0.10 & 0.20 & 0.275 & 0.30 & 0.35 \\
		\midrule
		0.60 & 86.50 & 86.50 & 86.52 & 86.52 & 86.43 \\
		0.75 & 86.38 & 86.38 & 86.47 & 86.64 & 86.42 \\
		0.80 & 86.61 & 86.61 & 86.59 & \textbf{86.68} & 86.43 \\
		0.85 & 86.63 & 86.63 & 86.50 & 86.66 & 86.44 \\
		\bottomrule
	\end{tabular}
	\label{tab:gate_grid_auc}
\end{table}

\begin{table}[t]
	\centering
	\caption{Effect of summary stride length in CEA.}
	\begin{tabular}{c|cc}
		\toprule
		\textbf{Summary Stride} & AUC (\%) w/o REA & AUC (\%) with REA \\
		\midrule
		3 & 86.69 & 87.90 \\
		5 & 86.68 & 87.63 \\
		7 & 86.72 & 87.83 \\
		9 & 86.65 & 87.79 \\
		\bottomrule
	\end{tabular}
	\label{tab:summaryStride}
\end{table}

\begin{table}[t]
	\centering
	\caption{\textbf{Effect of the interval selection parameter $k_{\max}$ in REA.} 
		Performance is stable across a wide range of values, with the best result obtained at $k_{\max}=6$.}
	\begin{tabular}{c|c}
		\toprule
		\textbf{$k_{\max}$} & AUC (\%) \\
		\midrule
		2  & 87.00 \\
		4  & 87.53 \\
		6  & \textbf{87.63} \\
		8  & 87.49 \\
		10 & 87.28 \\
		\bottomrule
	\end{tabular}
	\label{tab:Kmax}
\end{table}

\subsection{REA Threshold/Hyperparameter Analysis}

We analyze the effect of the interval selection parameter $k_{\max}$ used in REA, which controls the maximum number of anomaly intervals retained during dynamic interval selection. The results are reported in Table~\ref{tab:Kmax}. 

Overall, the method shows strong robustness across different values of $k_{\max}$, with performance remaining stable within a narrow range of AUC scores. The best performance is achieved at $k_{\max}=6$, which provides a good balance between capturing multiple anomaly intervals and avoiding overly fragmented predictions. 

Smaller values of $k_{\max}$ may miss secondary anomaly intervals, while excessively large values may introduce noisy intervals with weaker evidence. Based on this analysis, we use $k_{\max}=6$ as the default setting in all experiments.

\subsubsection{REA Keyword and Negation Sensitivity}
\label{app:more_results:keywords}

\begin{table}[tb]
	\caption{AUC (\%) of our full pipeline when varying the anomaly keyword/negation configuration used to compute textual evidence in RACA.}
	\label{tab:KeywordAnalysis}
	\centering
	\begin{tabular}{@{}l|c@{}}
		\toprule
		Configuration & AUC (\%)\\
		\midrule
		No keywords / negation cues & 87.36\\
		Our selected keyword and negation set & \textbf{87.63}\\
		Expanded keyword set & 87.34\\
		\bottomrule
	\end{tabular}
\end{table}

\begin{table}[t]
	\centering
	\begin{minipage}{0.62\linewidth}
		\centering
		\setlength{\tabcolsep}{4pt}
		\renewcommand{\arraystretch}{1.15}
		\caption{\textbf{Cue and negation coefficient analysis.} 
			Anomaly cue coefficient $\gamma$ is shown along the x-axis and negation penalty $\delta$ along the y-axis.}
		\small
		\begin{tabular}{c|ccccc}
			\toprule
			$\delta \downarrow \gamma \rightarrow$ & 0.03 & 0.05 & 0.07 & 0.09 & 0.11 \\
			\midrule
			0.15 & 87.27 & 87.27 & 87.27 & 87.27 & 87.27\\
			0.20 & 87.64 & 87.53 & 87.53 & 87.53 & 87.53\\
			0.25 & 87.64 & \textbf{87.63} & 87.63 & 87.63 & 87.52\\
			0.30 & 87.62 & 87.62 & 87.62 & 87.62 & 87.62\\
			0.35 & 87.21 & 87.17 & 87.17 & 87.17 & 87.17\\
			0.40 & 87.16 & 87.07 & 87.08 & 87.07 & 87.03\\
			\bottomrule
		\end{tabular}
		\label{tab:Negs_and_Cues_auc}
	\end{minipage}
	\hfill
	\begin{minipage}{0.35\linewidth}
		\centering
		\caption{\textbf{Sensitivity of $\alpha$.} 
			$\alpha$ controls the weight of the segment-level anomaly prediction in Eq.~(2).}
		\small
		\begin{tabular}{c|c}
			\toprule
			$\alpha$ & AUC (\%) \\
			\midrule
			0.80 & 87.15 \\
			0.85 & 87.63 \\
			0.90 & \textbf{87.63} \\
			0.95 & 87.55 \\
			1.00 & 87.28 \\
			\bottomrule
		\end{tabular}
		\label{tab:alpha_auc}
	\end{minipage}
\end{table}

REA incorporates lightweight textual signals through anomaly keyword matches and explicit negation suppression (Appendix~\ref{app:cue_words}). 
To verify robustness, we first ablate the keyword design by (i) removing keywords entirely and (ii) expanding the keyword list beyond our selected set. 
Results in Table~\ref{tab:KeywordAnalysis} show that the curated keyword set provides the best performance, while overly broad expansions slightly reduce AUC, likely due to spurious cue activations.

We further analyze the sensitivity of the weighting coefficients used in Eq.~(2) of the main paper. 
Table~\ref{tab:Negs_and_Cues_auc} evaluates the anomaly cue coefficient $\gamma$ and the negation penalty $\delta$. 
Performance remains highly stable across a broad range of settings, indicating that the REA evidence aggregation mechanism is not sensitive to precise coefficient tuning. 
Moderate negation penalties ($\delta \in [0.20,0.30]$) consistently produce the strongest results, while the cue keyword coefficient $\gamma$ has only a minor effect within the tested range.

We additionally study the influence of the segment-level anomaly weight $\alpha$, which balances the original prediction score with textual evidence. 
As shown in Table~\ref{tab:alpha_auc}, performance peaks for $\alpha \in [0.85,0.90]$ and remains stable across nearby values. 
This indicates that textual evidence provides a useful complementary signal while the original segment-level prediction remains the dominant component. 
Based on these results, we adopt $(\alpha,\gamma,\delta)=(0.90,0.05,0.25)$ for all experiments.

\subsection{Sensitivity Test on Vision Encoders}

We further evaluate the robustness of the proposed pipeline to the choice of vision encoder used for diverse key frame selection in CEA. In the official implementation, we use a pretrained ResNet-50 model to encode representative frames. We also replace this encoder with CLIP features to assess the sensitivity of the method to the visual representation.

As shown in Table~\ref{tab:Encoders}, the choice of encoder has minimal impact on performance. Using CLIP features results in only a slight decrease in AUC, indicating that the key frame selection stage is largely insensitive to the specific vision encoder used.

\begin{table}[tb]
	\caption{AUC (\%) when using different vision encoders for diverse key frame selection in CEA.}
	\label{tab:Encoders}
	\centering
	\begin{tabular}{@{}l|c@{}}
		\toprule
		Vision Encoder & AUC (\%)\\
		\midrule
		Pretrained ResNet-50 & 87.63\\
		CLIP & 87.54\\
		\bottomrule
	\end{tabular}
\end{table}

\begin{table}[tb]
	\centering
	\caption{mIoU (\%) of different components of \name \space compared to the baseline VERA.}
	\label{tab:miou}
	\begin{tabular}{l|c}
		\toprule
		Method & mIoU (\%)\\
		\midrule
		\multicolumn{2}{c}{\textit{Without REA}} \\
		\midrule
		Baseline VERA & 30.06\\
		Non-gated Summaries & 31.91\\
		Gated Summaries (CEA) & 28.94\\
		\midrule
		\multicolumn{2}{c}{\textit{With REA}} \\
		\midrule
		No Summaries + REA & 34.04\\
		Non-gated Summaries + REA & 35.12\\
		Gated Summaries + REA (\name) & \textbf{35.16}\\
		\bottomrule
	\end{tabular}
\end{table}

\subsection{Performance Comaprison on Other Metrics}
In addition to AUC, we evaluate our method using mean intersection-over-union (mIoU) to assess temporal localization quality. The results are reported in Table~\ref{tab:miou}. Overall, \name \space achieves the highest mIoU among all configurations.

Notably, REA contributes the largest improvement in localization performance. Applying REA alone increases mIoU from 30.06\% to 34.04\%, indicating that recursive temporal aggregation effectively refines anomaly boundaries. When combined with CEA, the performance further improves to 35.16\%.

We also observe that contextual summaries alone do not directly improve mIoU, as gated summaries without REA slightly decrease the score. This is expected since CEA primarily improves contextual reasoning for anomaly detection rather than boundary refinement. However, when combined with REA, the contextual signals help identify more coherent anomaly intervals, resulting in the best overall localization performance.

\section{LLM-as-a-Judge and Human Evaluation Protocol}
\label{app:llm_judge}

To quantify whether our event-level explanations are \emph{more descriptive}
than segment-level explanations, we evaluate how well an external LLM can
recover the ground-truth anomaly \emph{category} when given \textbf{only the explanation text}
(no frames, no scores, no metadata).
We implement this ``LLM-as-a-judge'' protocol using \texttt{gpt-oss:20b}.

\subsubsection{Closed-Set Category Space}
\label{app:prompts:llm_judge_labels}

The judge performs \emph{closed-set} classification over the 13 UCF-Crime anomaly types used in our UCF-Crime experiments:

\begin{verbatim}
	[
	"abuse", "arrest", "arson", "assault", "burglary", "explosion", 
	"fighting", "roadaccidents", "robbery", "shoplifting", 
	"shooting", "stealing", "vandalism"
	]
\end{verbatim}

\paragraph{Alias normalization.}
Because natural language explanations may use synonymous surface forms
(e.g., ``accident'' $\rightarrow$ \texttt{roadaccidents}, ``fight'' $\rightarrow$ \texttt{fighting}),
we map common aliases to the canonical labels using a deterministic lookup table
prior to computing accuracy.

\subsubsection{Ground-Truth Category Inference from Video Names}
\label{app:prompts:llm_judge_gold}

UCF-Crime video names encode the anomaly type (e.g., \texttt{Robbery102\_x264}).
We infer the ground-truth category by extracting the leading alphabetic prefix
before the first digit (with fallbacks for underscore formatting), then
canonicalizing via the alias table.
Videos whose names indicate ``Normal'' content are treated as out-of-scope
for this protocol and skipped.

\subsubsection{Explanation Variants Compared}
\label{app:prompts:llm_judge_variants}

For each predicted anomalous cluster, we compare four explanation types:
\begin{enumerate}
	\item \textbf{Cluster-Level:} the generated cluster explanation.
	\item \textbf{Most Anomalous Segment-Level (Peak):} the single segment-level explanation inside the cluster
	with highest local evidence score (Sec.~\ref{app:prompts:llm_judge_peak}).
	\item \textbf{Random Segment-Level:} a randomly selected segment-level explanation from within the cluster.
	\item \textbf{All Segment-Level Concatenated:} concatenation of all segment-level explanations within the cluster
	(optionally including per-segment running summaries).
\end{enumerate}
All variants are judged independently by the same LLM judge prompt and decoding.

\subsubsection{Peak Segment Selection (Most Anomalous Segment-Level Baseline)}
\label{app:prompts:llm_judge_peak}

To define the ``most anomalous'' segment-level baseline robustly, we select the peak segment
using the soft evidence scores introduced in Eq. (2) in the main paper. 

\subsubsection{Judge Prompt Template}
\label{app:prompts:llm_judge_prompt}

We prompt the external judge with a strict closed-set instruction and require
a single JSON output containing the predicted label.

\begin{figure}[t]
	\centering
 
		\begin{minipage}{0.95\linewidth}
			\small
			\textbf{CATEGORY\_JUDGE\_PROMPT (verbatim)}\\
			\begin{verbatim}
				You are a strict evaluator.
				
				Task: Given ONLY the text explanation of an event in a video, predict the video anomaly category.
				
				Closed-set labels (choose exactly ONE):
				{labels}
				
				Rules:
				- Use ONLY the information explicitly stated in the explanation.
				- Output must be a single JSON object with one key: "label".
				
				Explanation:
				{explanation}
				
				Return ONLY:
				{"label": "<one of the labels above>"}
			\end{verbatim}
		\end{minipage}
 
	\caption{Prompt used for explanation-only anomaly category prediction by the external LLM judge.}
	\label{fig:judge_prompt}
\end{figure}

\paragraph{Structured output constraint.}
We enforce JSON-schema constrained decoding with the enum:
\texttt{CANON\_CATS $\cup$ \{"unknown"\}}.
Any unparseable response, schema violation, or runtime error triggers up to
three retries; after failure, the prediction defaults to \texttt{unknown}.

\subsubsection{Decoding Configuration}
\label{app:prompts:llm_judge_decoding}

We evaluate with deterministic decoding to ensure stable judge behavior:
\begin{itemize}
	\item \textbf{Judge model:} \texttt{gpt-oss:20b}
	\item \textbf{Backend:} \texttt{ollama}
	\item \textbf{Temperature:} $0.0$
\end{itemize}

\subsubsection{Accuracy Metric}
\label{app:prompts:llm_judge_metric}

For each explanation variant, we compute \textbf{category recovery accuracy}:
\[
\mathrm{Acc} = \frac{1}{N}\sum_{n=1}^{N} \mathbb{I}\big[\hat{c}_n = c_n\big],
\]
where $c_n$ is the inferred ground-truth category from the video filename
(Sec.~\ref{app:prompts:llm_judge_gold}) and $\hat{c}_n$ is the judge prediction after canonicalization.
This yields the values reported in the reported results.

\subsection{Human Evaluation Protocol}

To assess the quality of the generated explanations, we conduct a human evaluation following the procedure described in the main paper. We randomly sample 200 predicted anomalous events from the UCF-Crime dataset and collect the explanations generated by our \name \space as well as 3 comparison segment-level explanations, which are the same as the LLM as a judge evaluation (see Appendix section \ref{app:prompts:llm_judge_variants}).

For each event, annotators are shown the representative frames corresponding to the detected anomaly together with the candidate explanations. The evaluation is conducted as a pairwise preference task: annotators are asked to select the explanation that better describes the anomalous event while remaining faithful to the visual evidence.

Annotators are instructed to prioritize factual consistency with the visual content and the completeness of the anomaly description rather than stylistic aspects of the text. Each comparison is evaluated independently, and the final human preference score is computed as the percentage of times an explanation is selected as the better description.

When combined with anomaly category prediction accuracy and average token length, these metrics provide a comprehensive evaluation of explanation usefulness, balancing human preference, semantic accuracy, and descriptive efficiency.

\section{Computational Complexity Analysis}
\label{app:complexity}

We analyze the computational complexity of the proposed Context-Enhanced Analysis (CEA) and Recursive Evidence Aggregation (REA), with particular emphasis on the additional cost of CEA relative to a VERA-style segment-level inference pipeline.

\paragraph{Notation.}
Let $h$ denote the number of temporal segments in a video. Let $C_{\text{vlm}}$ denote the cost of one segment-level VLM inference call, which is the dominant cost in both VERA and CEA. Let $C_{\text{sum}}$ denote the cost of one summary-generation call, $C_{\text{hist}}$ the cost of encoding one representative history frame, and $C_{\text{grd}}$ the cost of one grounding step. Let $s$ denote the summary refresh stride (with $s=5$ in our implementation), and let $M$ denote the maximum number of stored history snippets (with $M=50$ in our implementation).

\paragraph{VERA-style segment inference.}
A VERA-style pipeline performs one VLM inference per segment and does not maintain historical summaries or grounding statistics. Its total complexity is therefore
\[
O(h\,C_{\text{vlm}}).
\]

\paragraph{Analysis on CEA.}
CEA preserves the same dominant computation as VERA: it still performs exactly one segment-level VLM inference for each of the $h$ segments. Thus, the main computational term remains
\[
O(h\,C_{\text{vlm}}).
\]
Compared to VERA, the additional computation in CEA comes only from three lightweight components:

\begin{enumerate}
	\item \textbf{History update.} For each segment, CEA encodes one representative frame and inserts it into a bounded history buffer, contributing
	\[
	O(h\,C_{\text{hist}}).
	\]
	
	\item \textbf{Periodic summary generation.} A history summary is refreshed only once every $s$ segments, rather than at every segment. This contributes
	\[
	O\!\left(\frac{h}{s}\,C_{\text{sum}}\right).
	\]
	
	\item \textbf{Grounding and gating.} At the same refresh steps, CEA computes image-text similarities and entropy-based grounding statistics over a bounded set of history key frames, contributing
	\[
	O\!\left(\frac{h}{s}\,C_{\text{grd}}\right).
	\]
\end{enumerate}

Hence, the total complexity of CEA is
\[
O\!\left(
h\,C_{\text{vlm}}
+
h\,C_{\text{hist}}
+
\frac{h}{s}\,C_{\text{sum}}
+
\frac{h}{s}\,C_{\text{grd}}
\right).
\]

Importantly, this additional overhead is modest in practice. Unlike the main anomaly scoring step, which invokes the full VLM on every segment, summary generation and grounding are performed only periodically, and both operate on bounded history size and a small number of selected key frames. As a result, CEA does not alter the dominant linear scaling of VERA with respect to the number of segments. Instead, it adds only a small constant-factor overhead while providing temporally grounded contextual reasoning.

Equivalently, relative to VERA, the extra compute introduced by CEA is
\[
O\!\left(
h\,C_{\text{hist}}
+
\frac{h}{s}\,C_{\text{sum}}
+
\frac{h}{s}\,C_{\text{grd}}
\right),
\]
which is practically minor because $s$ is fixed, the history size is capped, and the dominant cost remains the per-segment VLM inference. Therefore, both VERA and CEA scale linearly with video length, but CEA achieves context-aware reasoning with only minimal additional computation.

\paragraph{Analysis on REA.}
REA is a lightweight post-processing step applied after segment-level predictions are obtained. Let $h$ again denote the number of temporal segments. The recursive binary partitioning evaluates temporal windows and splits only those that satisfy the anomaly evidence criterion. Since the recursion depth is bounded by $O(\log h)$ and the windows at each depth partition the temporal axis, the total amount of processing is $O(h\log h)$ in the worst case. Interval merging and top-$K$ interval selection add at most linearithmic overhead in the number of candidate intervals, which is upper bounded by $h$. Therefore, the overall worst-case complexity of REA is
\[
O(h \log h).
\]

In practice, the computational cost of REA is negligible compared to the VLM inference cost in CEA or VERA, since REA operates only on segment-level scores and short textual responses and does not invoke an additional large vision-language model.

\paragraph{Event-level explanation generation.}
After REA produces cleaned segment-level anomaly scores, we optionally generate event-level explanations by first finding the identified anomaly events and then invoking the VLM once per detected event. Let $c$ denote the number of anomaly events in a video. In practice, $c$ is small (typically around 2 per video), since anomalous events are sparse and temporally localized.

For each anomaly-event, the explanation module selects a bounded number of informative segments, extracts a fixed number of representative frames, and constructs a single prompt for event-level description. Since the number of selected segments, sampled frames, and generated tokens are all capped by fixed constants in our implementation, the cost per anomaly-event is bounded by a constant $C_{\text{evt}}$. The total complexity of event-level explanation generation is therefore
\[
O(c\,C_{\text{evt}}).
\]

Because $c \ll h$ in practice is typically very small, this cost is negligible compared to the segment-level VLM inference cost of CEA or VERA, both of which scale with the full number of temporal segments. Therefore, event-level explanation generation introduces only minimal additional computation while providing compact natural-language descriptions of the detected anomaly events.

\section{Discussion of Limitations and Potential Societal Impact}

\paragraph{Limitations.}
Despite its effectiveness in enabling structured temporal reasoning for VAD, the proposed framework has certain limitations that open avenues for future improvement. 
First, the method relies on the reasoning capabilities of large vision-language models (VLMs), whose outputs may occasionally contain hallucinations despite the grounding and evidence aggregation mechanisms used in CEA and REA. Second, the anomaly cue keywords and negation patterns used in REA are manually specified and may not cover all possible anomaly descriptions across diverse domains. 
Finally, the system inherits the computational cost of VLM inference, which may limit real-time deployment in resource-constrained environments.

\paragraph{Potential societal impact.}
Video anomaly detection systems are increasingly deployed in surveillance, security, and public safety contexts. By improving long-range temporal reasoning and producing more coherent event-level interpretations, our framework enhances the reliability and explainability of automated monitoring systems. Such improvements can assist human operators in identifying dangerous events more accurately, reducing fragmented alarms, and improving situational awareness in complex environments.

At the same time, video anomaly detection is inherently imperfect. Errors such as false alarms or missed detections cannot be fully eliminated and may have operational consequences in safety-critical scenarios. Importantly, our method is designed as a decision-support tool rather than a fully autonomous system in the real world, and it still requires human oversight for verification and final judgment. Responsible deployment should therefore ensure appropriate governance, transparency, and adherence to privacy regulations, with humans remaining in the loop for critical decisions.

\end{document}